\documentclass[11pt,a4paper]{article}
\usepackage[hyperref]{emnlp2020}
\usepackage{times}
\usepackage{latexsym}

\usepackage{microtype}
\usepackage{url}

\usepackage{booktabs}
\usepackage{multirow}
\usepackage{subcaption}

\usepackage{amsmath}

\renewcommand{\vec}[1]{\mathbf{#1}} 
\usepackage{color, colortbl}
\definecolor{Gray}{gray}{0.9}

\usepackage{CJKutf8}

\usepackage{tikz}
\usepackage{tikz-qtree}
\usetikzlibrary{arrows,decorations.pathmorphing,backgrounds,positioning,fit,petri,shapes.misc, arrows.meta,shapes.geometric,decorations.markings,calc,shadows.blur,decorations.pathreplacing,quotes}
\definecolor{myblue}{RGB}{6, 82, 221}
\definecolor{myorange}{RGB}{211, 84, 0}
\definecolor{lowblue}{RGB}{102,178,255}
\definecolor{justblue}{RGB}{84, 160, 255}
\definecolor{mypurple}{RGB}{108, 92, 231}
\definecolor{mygray}{RGB}{158, 158, 158}
\definecolor{lowpurple}{RGB}{204,153,255}
\definecolor{lowwhite}{RGB}{255,255,255}
\definecolor{verylowpurple}{RGB}{255,102,102}
\definecolor{embcolor}{RGB}{255,255,255}
\definecolor{myred}{RGB}{235, 47, 6} 
\definecolor{mygreen}{RGB}{162, 217, 206} 
\definecolor{fontgrey}{RGB}{44, 62, 80}
\definecolor{lowpurple}{RGB}{210, 180, 222}
\definecolor{mypumpkin}{RGB}{229, 152, 102}
\definecolor{lowgreen}{RGB}{171, 235, 198}
\definecolor{lowgreen2}{RGB}{186, 220, 88}
\definecolor{lowred}{RGB}{245, 183, 177}
\definecolor{lowyellow}{RGB}{241, 196, 15}
\definecolor{mypink}{RGB}{255, 118, 117}
\definecolor{bluemartina}{RGB}{18, 203, 196}
\definecolor{puffin}{RGB}{250, 152, 58}
\definecolor{grass}{RGB}{0, 148, 50}
\definecolor{cnngray}{RGB}{116, 125, 140}

\newcommand{\squishlist}{
	\begin{list}{$\bullet$}
		{ \setlength{\itemsep}{0pt}
			\setlength{\parsep}{3pt}
			\setlength{\topsep}{3pt}
			\setlength{\partopsep}{0pt}
			\setlength{\leftmargin}{1.5em}
			\setlength{\labelwidth}{1em}
			\setlength{\labelsep}{0.5em} } }

	\newcounter{Lcount}
	\newcommand{\squishlisttwo}{
		\begin{list}{\arabic{Lcount}. }
			{ \usecounter{Lcount}
				\setlength{\itemsep}{0pt}
				\setlength{\parsep}{0pt}
				\setlength{\topsep}{0pt}
				\setlength{\partopsep}{0pt}
				\setlength{\leftmargin}{2em}
				\setlength{\labelwidth}{1.5em}
				\setlength{\labelsep}{0.5em} } }
		
		\newcommand{\squishend}{
	\end{list} }

\usepackage{multirow}
\usepackage{hhline}
\usepackage{tabularx}
\usepackage{ragged2e}
\newcolumntype{Y}{>{\RaggedRight\let\newline\\\arraybackslash\hspace{0pt}}X} 
\usepackage{booktabs}
\usepackage{pgfplots}
\usepackage{pgfpages}
\usepackage{subcaption}
\usepackage{mathbbol}
\usepackage{CJKutf8}
\usepackage{arydshln}
\usepackage{amsmath}
\usepackage{graphicx}

\usepackage{amssymb}
\usepackage{amsfonts}

\usepackage{algorithm}
\usepackage{algorithmic}
\usepackage{pgf}

\usepackage{lipsum}
\usepackage{multicol}
\usepackage{changepage}

\usepackage[export]{adjustbox}

\usepackage[utf8]{inputenc}

\usepackage{amsthm}

\usepackage{flushend}
\renewcommand{\vec}[1]{\mathbf{#1}}

\aclfinalcopy 
\def\aclpaperid{1706} 

\title{ENT-DESC: Entity Description Generation by Exploring \\Knowledge Graph}

\author{ 
\textbf{
Liying Cheng\thanks{~~Liying Cheng is under the Joint Ph.D. Program between Alibaba and Singapore University of Technology and Design.} \textsuperscript{\rm ~1,2},
Dekun Wu\thanks{~~Dekun Wu was a visiting student at SUTD. Yan Zhang and Zhanming Jie were interns at Alibaba.}\textsuperscript{\rm ~~3},
Lidong Bing\textsuperscript{\rm 2},
Yan Zhang\footnotemark[2]\textsuperscript{\rm ~~1},
Zhanming Jie\footnotemark[2]\textsuperscript{\rm ~~1},
Wei Lu\textsuperscript{\rm 1},
Luo Si\textsuperscript{\rm 2}}\\
\textsuperscript{\rm 1} Singapore University of Technology and Design\\
\textsuperscript{\rm 2} DAMO Academy, Alibaba Group~~
\textsuperscript{\rm 3} York University, Canada \\
{\small \tt \{liying.cheng, l.bing, luo.si\}@alibaba-inc.com, 
jackwu@eecs.yorku.ca,} \\
{\small \tt \{yan\_zhang, zhanming\_jie\}@mymail.sutd.edu.sg, luwei@sutd.edu.sg}
}

\date{}

\begin{document}
\maketitle
\begin{abstract}
Previous works on knowledge-to-text generation take as input a few RDF triples or key-value pairs conveying the knowledge of some entities to generate a natural language description.
Existing datasets, such as W\textsc{iki}B\textsc{io}, WebNLG, and E2E, basically have a good alignment between an input triple/pair set and its output text. 
However, in practice, the input knowledge could be more than enough, since the output description may only cover the most significant knowledge. 
In this paper, we introduce a large-scale and challenging dataset to facilitate the study of such a practical scenario in KG-to-text.
Our dataset involves retrieving abundant knowledge of various types of main entities from a large knowledge graph (KG), which makes the current graph-to-sequence models severely suffer from the problems of information loss and parameter explosion while generating the descriptions.
We address these challenges by proposing a multi-graph structure that is able to represent the original graph information more comprehensively.
Furthermore, we also incorporate aggregation methods that learn to extract the rich graph information.
Extensive experiments demonstrate the effectiveness of our model architecture.
\footnote{Our code and data are available at \url{https://github.com/LiyingCheng95/EntityDescriptionGeneration}.}

\end{abstract}

\section{Introduction}


KG-to-text generation, automatically converting knowledge into comprehensive natural language, is an important task in natural language processing (NLP) and user interaction studies \cite{damljanovic2010natural}.
Specifically, the task takes as input some structured knowledge, such as resource description framework (RDF) triples of WebNLG~\cite{gardent2017webnlg}, key-value pairs of  W\textsc{iki}B\textsc{io}~\cite{lebret2016neural} and E2E~\cite{novikova2017e2e}, to generate natural text describing the input knowledge.
In essence, the task can be formulated as follows: given a main entity, its one-hop attributes/relations (e.g., W\textsc{iki}B\textsc{io} and E2E), and/or multi-hop relations (e.g., WebNLG), the goal is to generate a text description of the main entity describing its attributes and relations.
Note that these existing datasets basically have a good alignment between an input knowledge set and its output text. Obtaining such data with good alignment could be a laborious and expensive annotation process. 
More importantly, in practice, the knowledge regarding the main entity could be more than enough, and the description may only cover the most significant knowledge. Thereby, the generation model should have such differentiation capability.

\begin{figure}[t!]
    \begin{center}
    \leavevmode
    \includegraphics[width=0.48\textwidth]{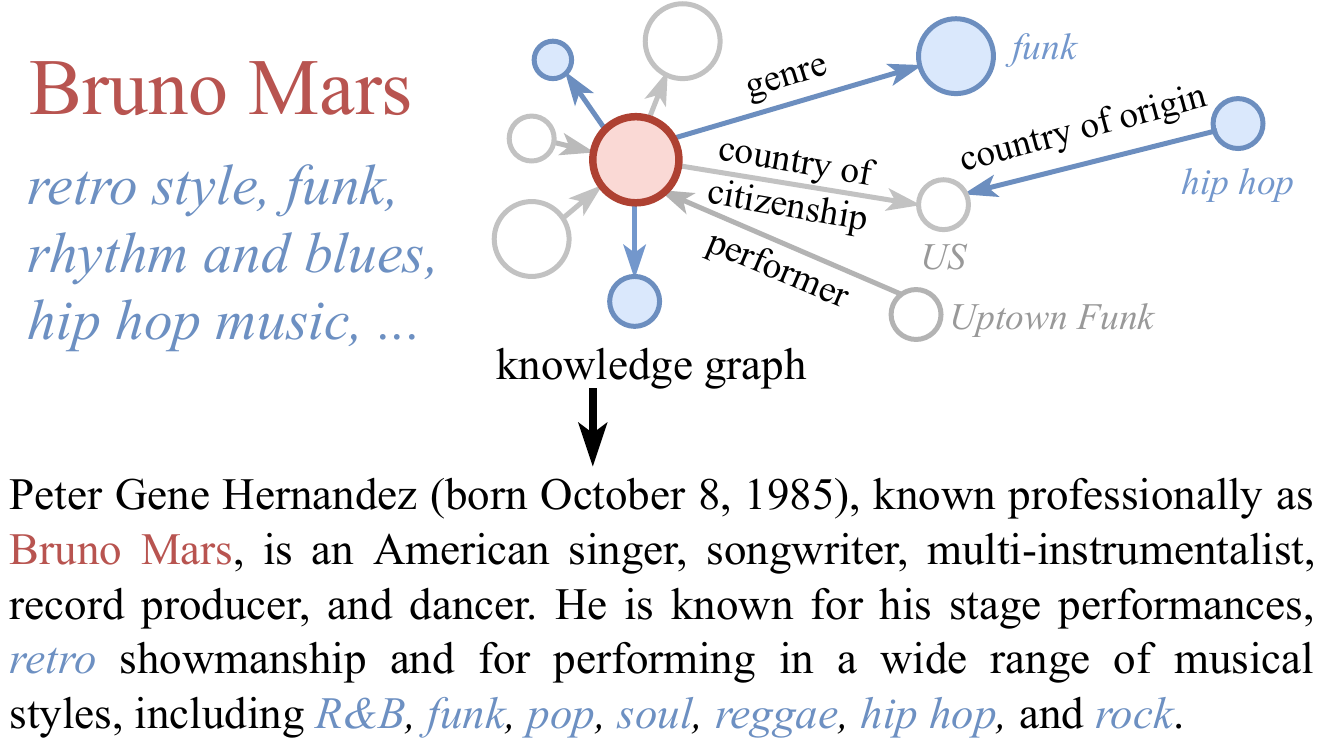}
    \caption{\label{eg} An example showing our proposed task.}
    \end{center}
\end{figure}


In this paper, we tackle 
an entity description generation task by exploring KG
in order to work towards more practical problems. 
Specifically, the aim is to generate a description with one or more sentences for a main entity and a few topic-related entities, which is empowered by the knowledge from a KG for a more natural description. 
In order to facilitate the study, we introduce a new dataset, namely \textit{entity-to-description} (ENT-DESC) extracted from Wikipedia and Wikidata, which contains over 110k instances.
Each sample is a triplet, containing a set of entities, the explored knowledge from a KG, and the description. 
Figure \ref{eg} shows an example to generate the description of the main entity, i.e., \textit{Bruno Mars}, given some relevant keywords, i.e., \textit{retro style, funk,} etc., which are called topic-related entities of \textit{Bruno Mars}. 
We intend to generate the short paragraph below to describe the main entity in compliance with the topic revealed by topic-related entities.
For generating accurate descriptions, one challenge is to extract the underlying relations between the main entity and keywords, as well as the peripheral information of the main entity.
In our dataset, we use such knowledge revealed in a KG, i.e., the upper right in Figure \ref{eg} with partially labeled triples. 
Therefore, to some extent, our dataset is a generalization of existing KG-to-text datasets.
The knowledge, in the form of triples, regarding the main entity and topic entities is automatically extracted from a KG, and such knowledge could be more than enough and not necessarily useful for generating the output. 

Our dataset is not only more practical but also more challenging due to lack of explicit alignment between the input and the output. Therefore, some knowledge is useful for generation, while others might be noise.
In such a case that many different relations from the KG are involved, standard graph-to-sequence models suffer from the problem of low training speed and parameter explosion, as edges are encoded in the form of parameters.
Previous work deals with this problem by transforming the original graphs into Levi graphs \cite{beck2018graph}.
However, Levi graph transformation only explicitly represents the relations between an original node and its neighbor edges, while the relations between two original nodes are learned implicitly through graph convolutional networks (GCN). 
Therefore, more GCN layers are required to capture such information \cite{marcheggiani2018deep}.
As more GCN layers are being stacked, it suffers from information loss from KG \cite{abu2018n}.
In order to address these limitations, we present a multi-graph convolutional networks (MGCN) architecture by introducing multi-graph transformation incorporated with an aggregation layer.
Multi-graph transformation is able to represent the original graph information more accurately, while the aggregation layer learns to extract useful information from the KG.
Extensive experiments are conducted on both our dataset and benchmark dataset (i.e., WebNLG).
MGCN outperforms several strong baselines, which demonstrates the effectiveness of our techniques, especially when using fewer GCN layers.

Our main contributions include: 
\squishlist
\item
We construct a large-scale dataset ENT-DESC for a more practical task of entity description generation by exploring KG.
To the best of our knowledge, ENT-DESC is the largest dataset of KG-to-text generation.
\item
We propose a multi-graph structure transformation approach that explicitly expresses a more comprehensive and more accurate graph information, in order to overcome limitations associated with Levi graphs.
\item
Experiments and analysis on our new dataset show that our proposed MGCN model incorporated with aggregation methods outperforms strong baselines by effectively capturing and aggregating multi-graph information.

 \squishend

\section{Related Work}


\paragraph{Dataset and Task.}
There is an increasing number of new datasets and tasks being proposed in recent years as more attention has been paid to data-to-text generation.
\citet{gardent2017webnlg} introduced the WebNLG challenge, which aimed to generate text from a small set of RDF knowledge triples (no more than 7) that are well-aligned with the text. To avoid the high cost of preparing such well-aligned data, researchers also studied how to leverage automatically obtained partially-aligned data in which some portion of the output text cannot be generated from the input triples \cite{fuzihao2020partially}.
\citet{koncel2019text} introduced AGENDA dataset, which aimed to generate paper abstract from a title and a small KG built by information extraction system on the abstracts and has at most 7 relations. 
In our work, we directly create a knowledge graph for the main entities and topic-related entities from Wikidata without looking at the relations in our output.
Scale-wise, our dataset consists of 110k instances while AGENDA is 40k.
\citet{lebret2016neural} introduced W\textsc{iki}B\textsc{io} dataset that generates the first sentence of biographical articles from the key-value pairs extracted from the article's infobox. 
\citet{novikova2017e2e} 
introduced E2E dataset in the restaurant domain, which aimed to generate restaurant recommendations given 3 to 8 slot-value pairs.
These two datasets were only for a single domain, while ours focuses on multiple domains of over 100 categories, including people, event, location, organization, etc. 
Another difference is that we intend to generate the first paragraph of each Wikipedia article from a more complicated KG, but not key-value pairs.
Another popular task is AMR-to-text generation \cite{konstas2017neural}. The structure of AMR graphs is rooted and denser, which is quite different from the KG-to-text task.
Researchers also studied how to generate texts from a few given entities or prompts \cite{DBLP:conf/aaai/LiBQC0Y19,DBLP:conf/aaai/FuBL20}. However, they did not explore the knowledge from a KG. 

\paragraph{Graph-to-sequence Modeling.}
In recent years, graph convolutional networks (GCN) have been applied to several tasks (e.g., semi-supervised node classification \citep{kipf2016semi}, semantic role labeling \citep{marcheggiani2017encoding} and neural machine translation \citep{bastings2017graph}) and also achieved state-of-the-art performance on graph-to-sequence modeling.
In order to capture more graphical information, ~\citet{velivckovic2017graph} introduced graph attention networks (GATs) through stacking a graph attentional layer, but only allowed to learn information from adjacent nodes implicitly without considering a more global contextualization.
~\citet{marcheggiani2017encoding} then used GCN as the encoder in order to capture more distant information in graphs.
Since there are usually a large amount of labels for edges in KG, such graph-to-sequence models without graph transformation will incur information loss and parameter explosion.
\citet{beck2018graph} proposed to transform the graph into Levi graph in order to work towards the aforementioned deficiencies, together with gated graph neural network (GGNN) to build graph representation for AMR-to-text problem. 
However, they face some new limitations brought in by Levi graph transformation: the entity-to-entity information is being ignored in Levi transformation, as also mentioned in their paper. Afterwards, deeper GCNs were stacked \cite{guo2019densely} to capture such ignored information implicitly.
In contrast, we intend to use fewer GCN layers to capture more global contextualization by explicitly stating all types of graph information with different transformations.

\section{Task Description}
In this paper, we tackle a practical problem of entity description generation by exploring KG.
In practice, it is difficult to describe an entity in only a few sentences as there are too many aspects for an entity.
Now, if we are given a few topic-related entities as topic restrictions to the main entity, the text to be generated could be more concrete, particularly when we are allowed to explore the connections among these entities in KG.
As seen in Figure \ref{eg}, when we are asked to use one or two sentences to introduce ``\textit{Bruno Mars}''\footnote{https://en.wikipedia.org/wiki/Bruno\_Mars}, his popular singles will first come into some people's minds, while his music genres might be in other people's first thought.
With the introduction of topic-related entities, the description will have some focus.
In this case, when topic-related entities, i.e., \textit{R\&B}, \textit{hip hop}, \textit{rock}, etc., are provided, we are aware of describing \textit{Bruno Mars} in the direction of music styles on top of their basic information.

Formally, given a set of entities $\vec{e} = \left \{ \textsc{e}_{1}, ..., \textsc{e}_{n} \right \}$ and a KG $\mathcal{G} = \left ( \mathcal{V}, \mathcal{E}\right )$, where $\textsc{e}_{1}$ is main entity, $\textsc{e}_{2}, ..., \textsc{e}_{n}$ are topic-related entities, $\mathcal{V}$ is the set of entity nodes and $\mathcal{E}$ is the set of directed relation edges.
We intend to generate a natural language text $\vec{y} = \{y_1, y_2,\cdots,y_{T}\}$.
Meanwhile, we explore $\mathcal{G}$ for useful information to allow a more natural description. Here, the KG $\mathcal{G}$ can also be written as a set of RDF triples: $\mathcal{G} = \left \{ \left < \textsc{v}_{\textsc{s}_{1}}, \textsc{p}_{1}, \textsc{v}_{\textsc{o}_{1}} \right >, ..., \left < \textsc{v}_{\textsc{s}_{\textsc{m}}}, \textsc{p}_{\textsc{m}}, \textsc{v}_{\textsc{o}_{\textsc{m}}} \right > \right \}$, where $\textsc{m}$ is the total number of triples, $\textsc{v}_{\textsc{s}_{i}}, \textsc{v}_{\textsc{o}_{i}} \in \mathcal{V}$ are the subject and object entities respectively, $\textsc{p}_{i}$ is the predicate stating the relation between $\textsc{v}_{\textsc{s}_{i}}$ and $\textsc{v}_{\textsc{o}_{i}}$.

\begin{table}[!t]
\centering
\scalebox{0.7}{
\setlength{\tabcolsep}{0.9mm}{
 \begin{tabular}{lcccc}
 \toprule
& \textbf{WebNLG}&\textbf{AGENDA} &\textbf{E2E}&\textbf{ENT-DESC}
\\ [0.5ex] 
 \midrule
 \# instances  & 43K & 41K & 51K & 110K\\
 Input vocab&4.4K& 54K  & 120 &420K\\
 Output vocab &7.8K& 78K &5.2K&248K\\
 \# distinct entities &3.1K&297K &77&691K\\
 \# distinct relations&358&7 &8&957\\
Avg. \# triples per input &3.0&4.4 &5.6&27.4\\
Avg. \# words per output &23.7&141.3 &20.3&31.0\\
  \bottomrule
 \end{tabular}}}
 \caption{Dataset statistics of WebNLG, AGENDA and our prepared ENT-DESC. 
 \label{tb:stats_ds}}
\end{table}


\begin{figure*}[!t]
    \centering
    \begin{subfigure}[t]{0.5\textwidth}
        \centering
        \scalebox{0.48}{
        \includegraphics[height=2.1in]{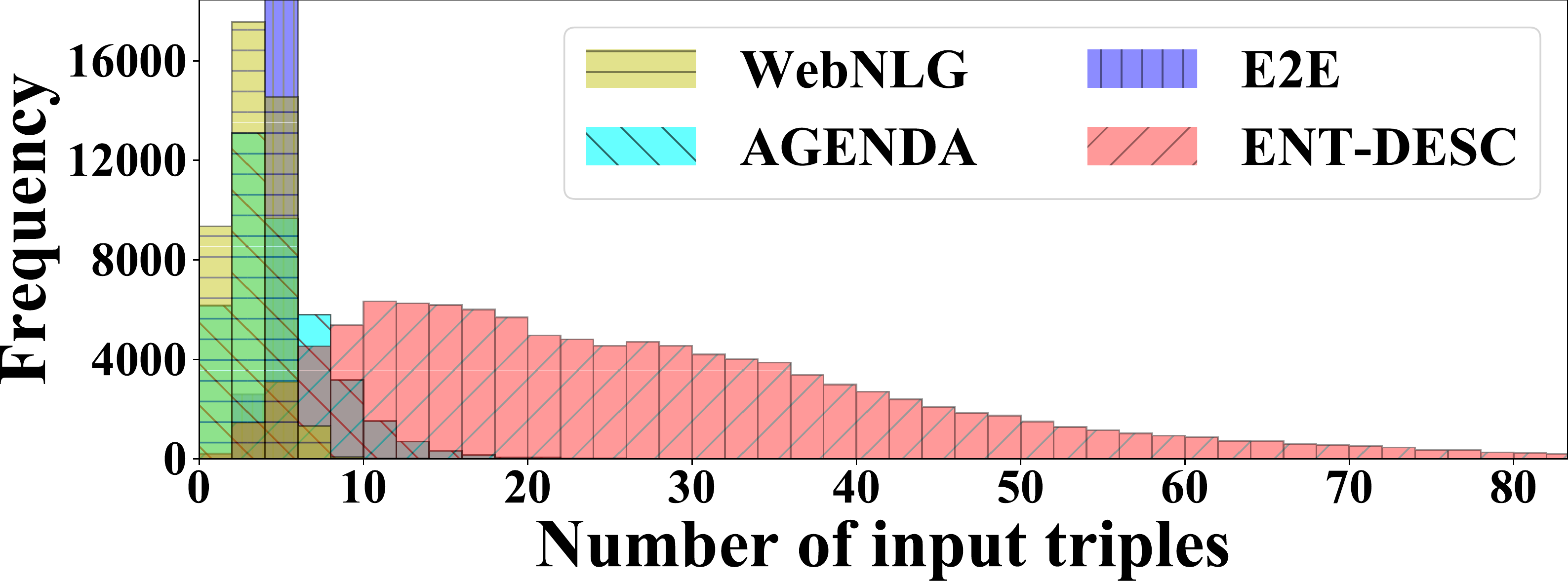}}
        \caption{Number of the input triples in each dataset.}
    \end{subfigure}%
    ~ 
    \begin{subfigure}[t]{0.5\textwidth}
        \centering
        \scalebox{0.48}{
        \includegraphics[height=2.1in]{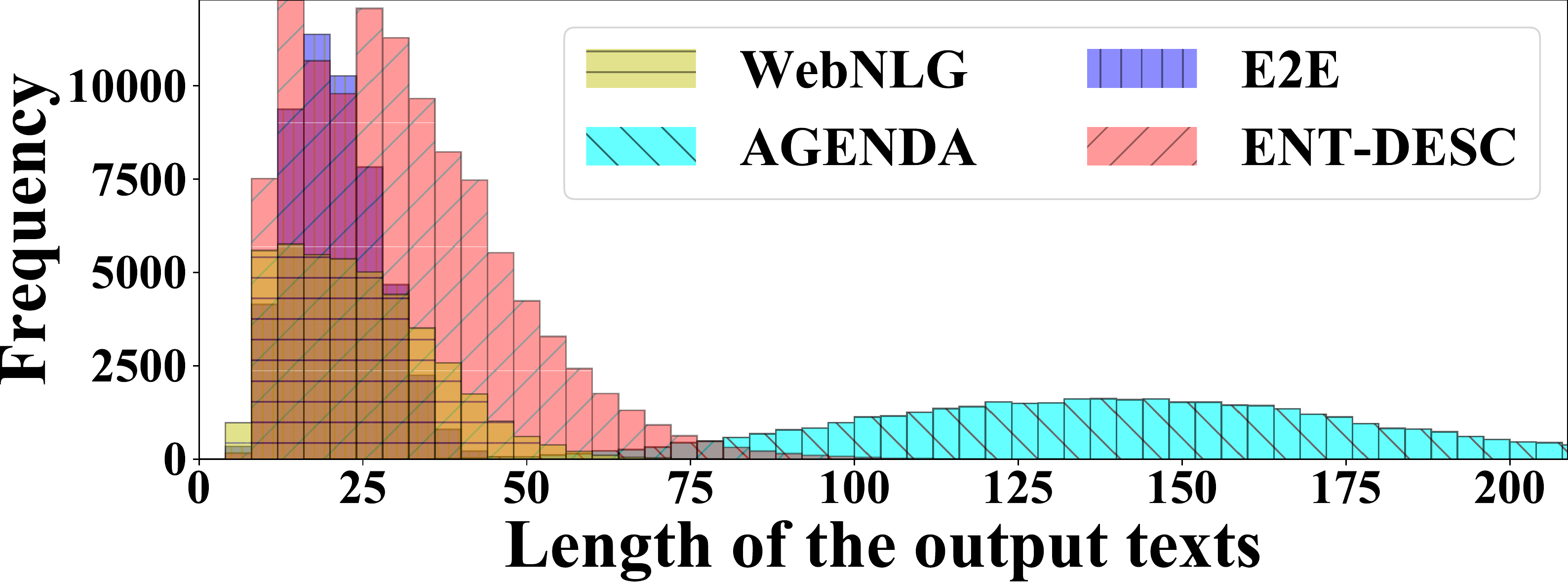}}
        \caption{Length of the output texts in each dataset.}
    \end{subfigure}
    \caption{Dataset comparison among WebNLG, AGENDA, E2E and our ENT-DESC.}
\label{fig:dataset_compa}
\end{figure*}

\section{ENT-DESC Dataset}



To prepare our dataset, we first use Nayuki's implementation\footnote{\url{https://www.nayuki.io/page/computing-wikipedias-internal-pageranks}} to calculate the PageRank score for more than 9.9 million Wikipedia pages.
We then extract the categories from Wikidata for the top 100k highest scored pages and manually select 90 categories out of the top 200 most frequent ones as the seed categories. 
The domains of the categories mainly include humans, events, locations and organizations.
The entities from these categories are collected as our candidate set of main entities.
We further process their associated Wikipedia pages for collecting the first paragraphs and entities with hyperlink as topic-related entities.
We then search Wikidata to gather neighbors of the main entities and 1-hop/2-hop paths between main entities and their associated topic-related entities, which finally results in a dataset consisting of more than 110k entity-text pairs with 3 million triples in the KG.
Although more-hop paths might be helpful, we limit to 1-hop/2-hop paths for the first study.
The comparison of our dataset with WebNLG, AGENDA and E2E is shown in Table~\ref{tb:stats_ds} and Figure~\ref{fig:dataset_compa}.

In the comparison of these four datasets, there are some obvious differences. First, our dataset is significantly larger than WebNLG, AGENDA and E2E (i.e., more than twice of their instances). 
Meanwhile, our vocabulary size and numbers of distinct entities/relations are all much larger. 
Second, the average number of input triples per instance is much larger than those of the other two. 
More importantly, our dataset provides a new genre of data for the task. Specifically, WebNLG has a strict alignment between input triples and output text, and accordingly, each input triple roughly corresponds to 8 words. 
AGENDA is different from WebNLG for generating much longer output, namely paper abstracts, with the paper title also given as input.
Moreover, as observed, quite a portion of text information cannot be directly covered by the input triples. E2E focuses on the restaurant domain with relatively simple inputs, including 77 entities and 8 relations in total.
Considering the construction details of these 3 datasets, all their input triples provide useful information (i.e., should be used) for generating the output. 
In contrast, our dataset has a much larger number of input triples, particularly considering the length difference of output texts. Lastly, another unique characteristic of our dataset is that not every input triple is useful for generation, which brings in the challenge that a model should be able to distill the helpful part for generating a better output sequence.

\section{Our MGCN Model}
\begin{figure*}[t!]
    \center{
    \scalebox{0.95}{
    \includegraphics[width=\linewidth
    ]
    {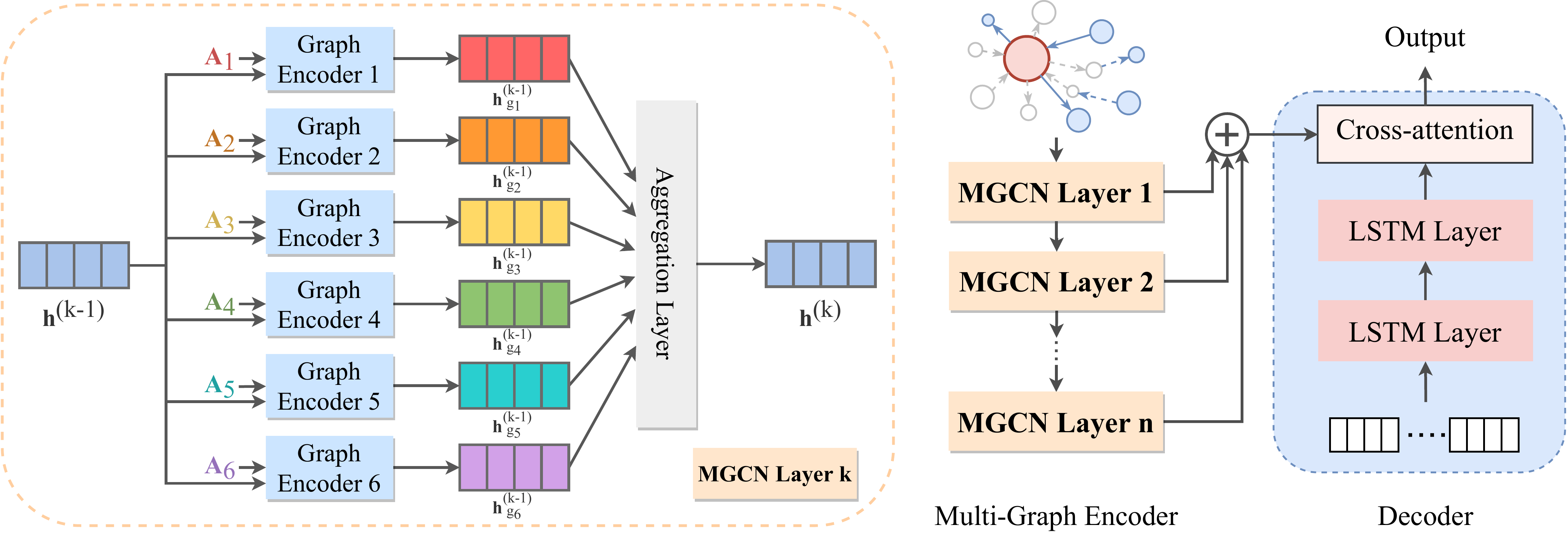}}}
    \caption{\label{layer} Overview of our model architecture. There are $n$ MGCN layers in the multi-graph encoder, and 2 LSTM layers in the decoder. $\vec{h}^{(k-1)}$ is the input graph representation at Layer $k$, and its 6 copies together with the corresponding adjacent matrices $\mathbf{A}_i$'s of transformed graphs in the multi graph (refer to Figure \ref{multigraph}) are fed into individual basic encoders.  Finally, we obtain the graph representation $\vec{h}^{(k)}$ for the next layer by aggregating the representations from these encoders.}
\end{figure*}

Given the explored knowledge, our task can be cast as a problem of generating text from KG.
We propose an encoder-decoder architecture with a multi-graph transformation, shown in Figure \ref{layer}.

\subsection{Multi-Graph Encoder}
We first briefly introduce the general flow of multi-graph encoder which consists of $n$ MGCN layers.
Before the first layer, graph embedding $\vec{h}^{(0)}$ representing a collection of node embeddings is initialized from input KG after multi-graph transformation.
By stacking $n$ MGCN layers accordingly with multi-graph transformation and aggregation, we obtain the final graph representation by aggregating the outputs of $n$ MGCN layers for decoding. We explain the details of an MGCN layer as follows. 

\paragraph{Graph Encoder.}
Before introducing our multi-graph transformation, we first look at our basic graph encoder in each MGCN layer (i.e., Graph Encoder 1 to 6 in Figure \ref{layer} left).
In this paper, we adopt graph convolutional networks (GCNs) \cite{duvenaud2015convolutional,kearnes2016molecular,kipf2016semi,marcheggiani2017encoding} as the basic encoder to consider the graph structure and to capture graph information for each node. More formally, given a directed graph $\mathcal{G^*} = \left ( \mathcal{V^*}, \mathcal{E^*}\right )$, we define a feature vector $\vec{x}_\textsc{v} \in \mathbb{R}^d$ for each node $\textsc{v} \in \mathcal{V^*}$. In order to capture the information of neighbors $\mathcal{N} (\cdot)$, the node representation $\vec{h}_{\textsc{v}_j}$ for each ${\textsc{v}_j} \in \mathcal{V}^*$ is calculated as:
\begin{center}
    \begin{tabular}{c}
		$
		\vec{h}_{\textsc{v}_j} =
		ReLU \Big (
		\sum\limits_{{\textsc{v}_i} \in \mathcal{N} ({\textsc{v}_j})}
		W_{\textsc{p}(i,j)} \vec{x}_{\textsc{v}_i}
		+
		\vec{b}_{\textsc{p}(i,j)}
		\Big ),
		$
	\end{tabular}
\end{center}
where $\textsc{p}(i,j)$ denotes the edge between node $\textsc{v}_i$ and $\textsc{v}_j$ including three possible directions: (1) $\textsc{v}_i$ to $\textsc{v}_j$, (2) $\textsc{v}_j$ to $\textsc{v}_i$, (3) $\textsc{v}_i$ to itself when $i$ equals to $j$. 
Weight matrix $W \in \mathbb{R}^{d \times d}$ and bias $\vec{b} \in \mathbb{R}^d$ are model parameters. $ReLU$ is the rectifier linear unit function. 
Only immediate neighbors of each node are involved in the equation above as it represents a single-layer GCN.

\begin{figure*}[t!]
    \center{
    \includegraphics[width=\linewidth]
    {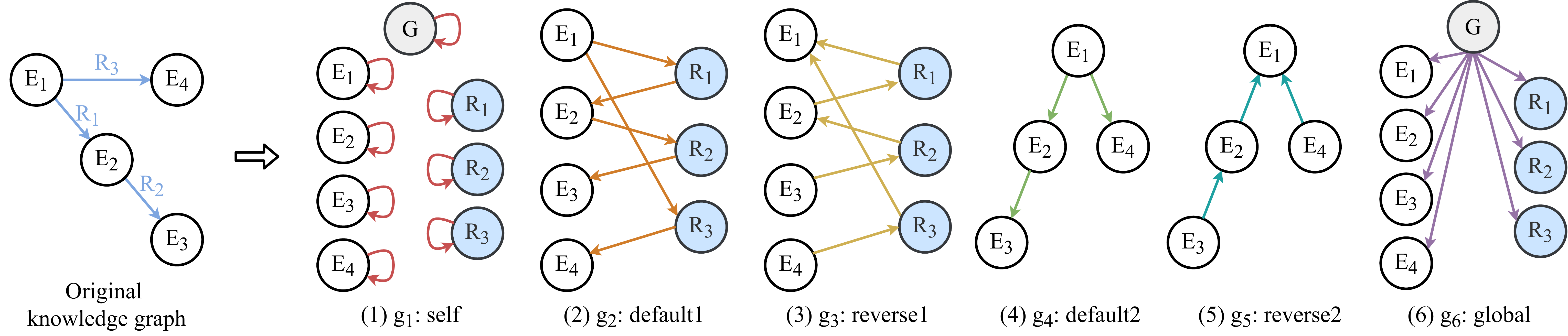}}
    \caption{\label{multigraph} An example of multi-graph transformation.
    }
\end{figure*}

\paragraph{Multi-Graph Transformation.}
\label{sec:mgt}
The basic graph encoder with GCN architecture as described above struggles with the problem of parameter explosion and information loss, as the edges are encoded in the form of parameters.
Previous works \cite{beck2018graph,guo2019densely,koncel2019text} deal with this deficiency by transforming the graph into a Levi graph.
However, Levi graph transformation also has its limitations, where entity-to-entity information is learned implicitly.
In order to overcome all the difficulties, we introduce a multi-graph structure transformation.
A simple example is shown in Figure \ref{multigraph}.
Given such a directed graph, where $\textsc{e}_{1}, \textsc{e}_{2}, \textsc{e}_{3}, \textsc{e}_{4}$ represent entities and $\textsc{r}_{1}, \textsc{r}_{2}, \textsc{r}_{3}$ represent relations in the KG, we intend to transform it into multiple graphs which capture different types of information. 
Similar to Levi graph transformation, all the entities and relations are represented as nodes in our multi-graph structure.
By doing such transformation, we are able to represent relations in the same format as entities using embeddings directly, which avoids the risk of parameter explosion.
This multi-graph transformation can be generalised for any graph regardless of the complexity and characteristic of the KG, and the transformed graph can be applied to any model architecture.

In this work, we employ a six-graph structure for our multi-graph transformation as shown in Figure \ref{multigraph}.
Firstly, in self graph (1), each node is assigned a self-loop edge namely \textit{self} label.
Secondly, graphs (2) and (3) are formed by connecting the nodes representing the entities and their adjacent relations.
In addition to connecting them in their original direction using \textit{default1} label, we also add a \textit{reverse1} label for the inverse direction of their original relations.
Thirdly, we create graphs (4) and (5) by connecting the nodes representing adjacent entities in the input graph, labeled by \textit{default2} and \textit{reverse2}, respectively.
These two graphs overcome the deficiency of Levi graph transformation by explicitly representing the entity-to-entity information from the input graph.
It also allows us to differentiate entities and relations by adding edges between entities.
Finally, in order to consider more global contextualization, we add a global node on top of the graph structure to form graph (6).
Each node is assigned with a \textit{global} edge directed from global node.
In the end, the set of transformed graphs can be represented by their edge labels $\mathcal{T} = \{ \textit{self}, \textit{default}, \textit{reverse}, \textit{default2}, \textit{reverse2}, \textit{global} \}$.

Given the six transformed graphs mentioned above, we construct six corresponding adjacency matrices: $\{\mathbf{A}_1, \mathbf{A}_2, \cdots, \mathbf{A}_6\}$. 
As shown in Figure \ref{layer} (left), these adjacency matrices are used by six basic graph encoders to obtain the corresponding transformed graph representations (i.e., $\vec{h}_g$).


\paragraph{Aggregation Layer.}
After learning 6 embeddings of multi graphs from the basic encoders at the current MGCN layer $k-1$, the model goes through an aggregation layer to obtain the graph embedding for the next MGCN layer $k$. 
We can get it by simply concatenating all 6 transformed graph embeddings with different types of edges.
However, such simple concatenation of the transformed graphs involves too many features and parameters.
In order to address the challenge mentioned above, 
we propose three aggregation methods for the multi-graph structure: sum-based, average-based and CNN-based aggregation.


Firstly, in sum-based aggregation layer, we compute the representation $\vec{h}^{(k)}$ at $k$-th layer as:
\begin{center}
    \begin{tabular}{c}
		$
		\vec{h}^{(k)} =
		\sum_{{g_i} \in {\mathcal{T}}}
		\vec{h}_{g_i}^{(k-1)},
		$
	\end{tabular}
\end{center}
where $\vec{h}_{g_i}^{(k-1)}$ represents the $i$-th graph representation, and $\mathcal{T}$ is the set of all transformed graphs. 
Sum-based aggregation allows a linear approximation of spectral graph convolutions and helps to reduce data sparsity and over-fitting problems.

Similarly, we apply an average-based aggregation method by normalizing each graph through a mean operation:
\begin{center}
    \begin{tabular}{c}
		$
		\vec{h}^{(k)} =
		\frac{1}{m}
		\sum_{{g_i} \in {\mathcal{T}}}
		\vec{h}_{g_i}^{(k-1)},
		$
	\end{tabular}
\end{center}
where $m$ is the number of graphs in $\mathcal{T}$.

We also try to employ a more complex CNN-based aggregation method. Formally, the representation $\vec{h}^{(k)}$ at $k$-th layer is defined as:
\begin{center}
    \begin{tabular}{c}
		$
		\vec{h}^{(k)} =
		W_{conv}
		\vec{h}_{mg}^{(k-1)}
        +
        \vec{b}_{mg}^{(k)}
		$
	\end{tabular}.
\end{center}

Here, we use convolutional neural networks (CNN) to convolute the multi-graph representation, where $\vec{h}_{mg} = \lbrack \vec{h}_{g_{1}}, ..., \vec{h}_{g_{6}} \rbrack$ is the representation of multi-graph and $\vec{b}_{mg}^{(k)}$ is the bias term.

By applying these aggregation methods, we obtain the graph representation for the next layer $\vec{h}^{(k)}$, which is able to capture different aspects of graph information more effectively by learning different types of edges in each transformed graph.

\paragraph{Stacking MGCN Layers.}
With the introduction of MGCN layer as described above, we can capture the information of higher-degree neighbors by stacking multiple MGCN layers.
Inspired by \citet{xu2018representation}, we employ a concatenation operation over $\vec{h}^{(1)}, \cdots, \vec{h}^{(n)}$ to aggregate the graph representations from all MGCN layers (Figure \ref{layer} right) to form the final layer $\vec{h}^{(final)}$, which can be written as follows:
\begin{center}
    \begin{tabular}{c}
		$
		\vec{h}^{(final)} =
		\left [
		\vec{h}^{(1)}, 
		\cdots 
		\vec{h}^{(n)}
		\right ].
        $
	\end{tabular}
\end{center}

Such a mechanism allows weight sharing across graph nodes, which helps to reduce overfitting problems.
To further reduce the number of parameters and overfitting problems, we apply the softmax weight tying technique \cite{press2017using} by tying source embeddings and target embeddings with a target softmax weight matrix.

\subsection{Attention-based LSTM Decoder}
We adopt the commonly-used standard attention-based LSTM as our decoder, where each next word $y_t$ is generated by conditioning on the final graph representation $\vec{h}^{(final)}$ and all words that have been predicted $y_1, ..., y_{t-1}$. The training objective is to minimize the negative conditional log-likelihood. 
Thus, the objective function can be written as:
\begin{center}
    \begin{tabular}{c}
		$
		\mathcal{L} =
        -
        \sum\limits_{t=1 }^{T} 
        \log p_{\boldsymbol{\theta}}(y_t | y_1, ..., y_{t-1}, \vec{h}^{(final)}),
		$
	\end{tabular}
\end{center}
where $T$ represents the length of the output sequence, and $p$ is the probability of decoding each word $y_t$ parameterized by $\boldsymbol{\theta}$.
As shown in the decoder from Figure \ref{layer}, we stack 2 LSTM layers and apply a cross-attention mechanism in our decoder.

\section{Experiments}

\begin{table*}[t!]
	\centering
	\scalebox{0.7}{
	    \setlength{\tabcolsep}{1.5mm}{
        \begin{tabular}{lccccccc}
        \toprule
        \bf Models & \bf BLEU & \bf METEOR & \bf TER$\boldsymbol{\downarrow}$  & \bf ROUGE$_1$ & \bf ROUGE$_2$ & \bf ROUGE$_L$ & \bf PARENT \\

        \midrule
        S2S~\cite{bahdanau2014neural} & \textcolor{white}{0}6.8 & 10.8 & 80.9 & 38.1 & 21.5 & 40.7 & 10.0 \\
        GraphTransformer \cite{koncel2019text} & 19.1 & 16.1 & 94.5 &  53.7 & 37.6 & 54.3 & 21.4 \\
        GRN \cite{beck2018graph} & 24.4 & 18.9 & 70.8 & 54.1 & 38.3 & 55.5 & 21.3\\
        GCN \cite{marcheggiani2018deep} & 24.8 & 19.3 & 70.4 & 54.9 & 39.1 & 56.2 & 21.8\\
        DeepGCN \cite{guo2019densely} & 24.9 & 19.3 & 70.2 & 55.0 & 39.3 & 56.2 & 21.8\\
        MGCN & \bf 25.7 & \bf 19.8 & \bf 69.3 &  \bf 55.8 & \bf 40.0 & \bf 57.0 & \bf 23.5\\
        \midrule
        MGCN + CNN & \bf 26.4 & \bf 20.4 & 69.4 &  56.4 &  40.5 & \bf 57.4 & \bf 24.2\\
        MGCN + AVG & 26.1 & 20.2 & \bf 69.2 & 56.4 & 40.3 & 57.3 & 23.9\\
        MGCN + SUM & \bf 26.4 & 20.3 & 69.8 & 56.4 &  \bf 40.6 &  \bf 57.4 & 23.9\\
        \midrule
        GCN + delex & 28.4 & 22.9 & 65.9 & 61.8 & 45.5 & 62.1 & 30.2\\
        MGCN + CNN + delex & 29.6 & \bf 23.7 & \bf 63.2 & \bf 63.0 & \bf 46.7 & \bf 63.2 & \bf 31.9\\
        MGCN + SUM + delex & \bf 30.0 & \bf 23.7 & 67.4 & 62.6 & 46.3 & 62.7 & 31.5\\
        \midrule \bottomrule \hline
        \multicolumn{8}{c}{The rows below are results of generating from entities only without exploring the KG.} \\
        \hline
        \rowcolor{Gray} E2S & 23.3 & 20.4 & 68.7 & 58.8 & 41.9 & 58.2 & 27.7 \\
        \rowcolor{Gray} E2S + delex & 21.8 & 20.5 & 67.5 & 59.5 & 39.5 & 59.2 & 23.4 \\
        \rowcolor{Gray} E2S-MEF & 24.2 & 21.3 & 65.8 & 59.8 & 43.3 & 60.0 & 26.3 \\
        \rowcolor{Gray} E2S-MEF + delex & 20.6 & 20.3 & 66.5 & 59.1 & 40.0 & 59.3 & 24.3 \\
        \hline
        \end{tabular}}}
    \caption{Main results of models on ENT-DESC dataset. $\boldsymbol{\downarrow}$ indicates lower is better.}
	\label{tab:main}
\end{table*}

\subsection{Experimental Settings}
We implement our MGCN architecture based on MXNET~\cite{chen2015mxnet} and Sockeye toolkit.
Hidden units and embedding dimensions for both encoder and decoder are fixed at 360. 
We use Adam \cite{kingma2014adam} with an initial learning rate of 0.0003 and update parameters with a batch size of 16. 
The training phase is stopped when detecting the convergence of perplexity on the validation set.
During decoding, we use beam search with a beam size of 10.
All models are run with V100 GPU.

We evaluate our models by applying both automatic and human evaluations.
For automatic evaluation, we use several common evaluation metrics: BLEU \cite{papineni2002bleu}, 
METEOR \cite{denkowski2011meteor}, TER \cite{snover2006study}, $\textsc{ROUGE}_1$, $\textsc{ROUGE}_2$, $\textsc{ROUGE}_L$ \cite{lin2004rouge}, PARENT \cite{dhingra2019handling}. We adapt MultEval \cite{clark2011better} and Py-rouge for resampling and significance test.



\subsection{Main Experimental Results}
We present our main experiments on ENT-DESC dataset and compare our proposed MGCN models with various aggregation methods against several strong GNN baselines \cite{bahdanau2014neural},  
GraphTransformer \cite{koncel2019text}, GRN \cite{beck2018graph}, GCN \cite{marcheggiani2018deep} and DeepGCN \cite{guo2019densely}, as well as a sequence-to-sequence (S2S) baseline.
We re-implement GRN, GCN and DeepGCN using MXNET.
We re-arrange the order of input triples following the occurrence of entities in output for S2S model to ease its limitation of not able to capture the graph structure.
We also apply sequence-to-sequence models on generating outputs directly from entities without exploring KG by (1) randomly shuffling the order of all input entities (E2S) and (2) randomly shuffling the order of all topic-related entities while keeping the \textbf{M}ain \textbf{E}ntity at \textbf{F}ront (E2S-MEF).
Furthermore, we apply a delexicalization technique on our dataset.
We delexicalize the main entity and topic-related entities by replacing these entities with tokens indicating the entity types and indices. 

Main results on our ENT-DESC dataset are shown in Table \ref{tab:main}.
Here, the numbers of layers in all baseline models and our MGCN models are set to be 6 for fair comparisons.
Our models consistently outperform the baseline models on all evaluation metrics.
S2S model has poor performance, mainly because the structure of our input triples is complicated as explained earlier.
Compared to GRN and GCN models, the BLEU score of MGCN model increases by 1.3 and 0.9, respectively. 
This result suggests the effectiveness of multi-graph transformation, which is able to capture more comprehensive information compared to the Levi graph transformation used by GCN and GRN (especially entity-to-entity information in the original graph). 
We then apply multiple methods of aggregation on top of the multi-graph structure.
MGCN+CNN and MGCN+SUM report the highest BLEU score of 26.4, followed by MGCN+AVG.
By applying our delexicalization technique, the results are further boosted by 3.2 to 3.6 BLEU scores for both baseline and our proposed models.
Moreover, our MGCN models and most baseline models outperform E2S and E2S-MEF, suggesting the importance of exploring KG when generating entity descriptions.
Compared to E2S and E2S-MEF, there is no further improvement after applying delexicalization (i.e., E2S+delex and E2S-MEF+delex). We speculate it is because the copy mechanism is incorporated in the sequence-to-sequence model. Some useful information in original entities may be lost when further applying the delexicalization.

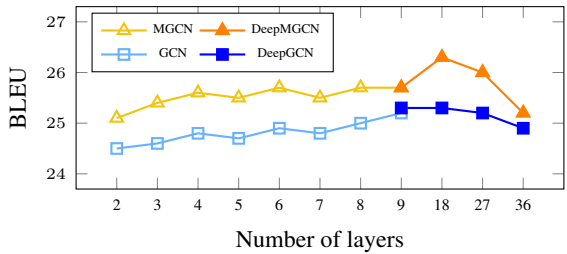
\begin{figure}
\begin{tikzpicture}
\pgfplotsset{width=8cm,height=4cm,compat=1.8}
\begin{axis}[
    xtick={1,2,3,4,5,6,7,8,9,10,11},
    ymin=24, ymax=27,
    xticklabels = {2,3,4,5,6,7,8,9,18,27,36},
    xticklabel style = {font=\fontsize{6}{1}\selectfont},
    yticklabel style = {font=\fontsize{6}{1}\selectfont},
    legend style={font=\fontsize{5}{1}\selectfont},
	ylabel={\footnotesize BLEU},
	xlabel={\footnotesize Number of layers},
	enlargelimits=0.1,
	legend style={at={(0.28,0.64)},anchor=south,legend columns=2}, 
	every axis plot/.append style={thick},
	tick label style={/pgf/number format/fixed},
    every node near coord/.append style={font=\tiny}
]

 \addplot[lowyellow] [mark=triangle,mark size=2.7pt] coordinates {
(1, 25.1) (2, 25.4) (3, 25.6) (4, 25.5) (5, 25.7) (6, 25.5) (7, 25.7) (8, 25.7)
 };
 
 \addplot[orange] [mark=triangle*,mark size=2.7pt] coordinates {
(8, 25.7) (9, 26.3) (10, 26.0) (11, 25.2) 
};
\addplot[lowblue]  [mark=square]  coordinates {
(1, 24.5) (2, 24.6) (3, 24.8) (4, 24.7) (5, 24.9) (6, 24.8) (7, 25.0) (8, 25.2)
};

\addplot[blue] [mark=square*]  coordinates {
(8, 25.3) (9, 25.3) (10, 25.2) (11, 24.9)
 };

\legend{ {MGCN}, {DeepMGCN}, {GCN} ,  {DeepGCN}}
\end{axis}
\end{tikzpicture}
\caption{Effect of different numbers of layers.}
\label{fig:layer_res}
\end{figure}

\subsection{Analysis and Discussion}

\paragraph{Effect of different numbers of MGCN layers.}
In order to examine the robustness of our MGCN models, we conduct further experiments by using different numbers of MGCN layers.
The results are shown in Figure \ref{fig:layer_res}.
We use MGCN to compare with the strongest baseline models using GCN according to the results in Table \ref{tab:main}.
More specifically, we compare to GCN on 2 to 9 layers and DeepGCN on 9, 18, 27 and 36 layers.
As shown in Figure \ref{fig:layer_res},
both models perform better initially as more GCN/MGCN layers are being stacked and start to drop afterward.
In general, MGCN/DeepMGCN achieves decent performance improvements of 0.3 to 1.0 from 2 to 36 layers, as shown in the line chart.
DeepMGCN achieves 26.3 BLEU score at 18 MGCN layers, which is 1.0 higher than deepGCN. 
It shows that, compared with learning the information implicitly by Levi graph, our multi-graph transformation brings in robust improvements by explicitly representing all types of information in the graph.
Another observation is that the BLEU score of MGCN with 3 layers (25.4) is already higher than the best performance of GCN/deepGCN.



\paragraph{Effect of various numbers of input triples.}

\begin{table}[t!]
	\centering
	\scalebox{0.7}{
	    \setlength{\tabcolsep}{1mm}{
\begin{tabular}{lcccc}
\toprule
\bf \# Input triples & \bf \# Instances & \bf GCN & \bf MGCN+SUM &  \bf $\Delta$ (BLEU) \\
\midrule
1 to 10 & 1,790 & 19.4 & 21.3 & +1.9 \\
11 to 20 & 2,999 & 22.6 & 24.6 & +2.0 \\
21 to 30 & 2,249 & 23.2 & 25.0 & +1.8 \\
31 to 50 & 2,830 & 31.6 & 32.8 & +1.2 \\
51 to 100 & 1,213 & 23.9 & 24.7 & +0.8 \\
\bottomrule
\end{tabular}}
    }
    \caption{Effect of different numbers of input triples.}
	\label{tab:triples}
\end{table}

\begin{table}[t!]
	\centering
	\scalebox{0.7}{
	    \setlength{\tabcolsep}{3mm}{
\begin{tabular}{lcc}
\toprule
\bf Model & \bf BLEU & \bf $\Delta$ (BLEU) \\
\midrule
MGCN + SUM & 26.4 & - \\
~~-- $g_6$: \textit{global} & 26.0 & -0.4 \\
~~-- $g_5$: \textit{reverse2} & 25.8 & -0.6 \\
~~-- $g_4$: \textit{default2} & 26.1 & -0.3 \\
~~-- $g_3$: \textit{reverse1} & 25.7 & -0.7 \\
~~-- $g_2$: \textit{default1} & 26.1 & -0.3 \\
\midrule
MGCN & 25.7 & -0.7 \\
GCN & 24.8 & -1.4 \\
\bottomrule
\end{tabular}}}
    \caption{Results of the ablation study.}
	\label{tab:ablation}
\end{table}

In order to have a deeper understanding of how multi-graph transformation helps the generation, we further explore the model performance under different numbers of triples on the test set.
Table \ref{tab:triples} shows the BLEU comparison between MGCN+SUM and GCN when using 6 layers.
Both models perform the best when the number of triples is between 31 and 50.
They both have a poorer performance when the number of triples is too small or too large, which should be due to the fact that the models have insufficient or very noisy input information for generation. 
Another observation is that the improvement of BLEU ($\Delta$) by our model is greater with a smaller number of input triples. It is plausibly because when the graph is larger, although our transformation techniques still bring in overall BLEU improvements, the increased graph complexity due to the transformation also hinders the generation.

\paragraph{Ablation Study.}
To examine the impact of each graph in our multi-graph structure, 
we show the ablation study in Table \ref{tab:ablation}.
Each transformed graph is removed respectively from MGCN+SUM with 6 layers, except for the $g_1$ (\textit{self}), which is always enforced in the graph \cite{kipf2016semi}.
We notice that the result drops after removing any transformed graph from the multi-graph.
Particularly, we observe the importance of $\{\textit{default2}, \textit{reverse2}\}$ and $\{\textit{default1}, \textit{reverse1}\}$ are equivalent, as the BLEU scores after removing them individually are almost the same.
This explains how multi-graph structure addresses the deficiency of Levi graph, i.e., entity-to-entity information 
is not represented explicitly in Levi graph. 
Additionally from the results, it is beneficial to represent the edges in the reverse direction for more effective information extraction in directed graphs as there are relatively larger gaps in BLEU drop after removing $g_3$ (\textit{reverse1}) or $g_5$ (\textit{reverse2}).

\begin{table}[t!]
	\centering
	\scalebox{0.7}{
	    \setlength{\tabcolsep}{2mm}{
\begin{tabular}{p{1cm}p{9cm}}
\toprule
Gold & The \colorbox{pink}{New Jersey Symphony Orchestra} is an American symphony \colorbox{lowblue}{orchestra} based in the state of \colorbox{lowblue}{New Jersey}. The NJSO is the state orchestra of New Jersey, performing concert series in six venues across the state, and is the resident orchestra of the \colorbox{lowblue}{New Jersey Performing Arts Center} in \colorbox{lowblue}{Newark, New Jersey}.\\
\midrule
GCN & The Newark Philharmonic Orchestra is an American orchestra based in \colorbox{lowblue}{Newark, New Jersey}, United States. \\\midrule
MGCN +SUM & The \colorbox{pink}{New Jersey Symphony Orchestra} is an American chamber \colorbox{lowblue}{orchestra} based in \colorbox{lowblue}{Newark, New Jersey}. The orchestra performs at the Newark Symphony Center at the Newark Symphony Center in \colorbox{lowblue}{Newark, New Jersey}.\\
\bottomrule
\end{tabular}}}
    \caption{An example of generated sentences.}
	\label{tab:case}
\end{table}



\paragraph{Case Study.}
Table \ref{tab:case} shows example outputs generated by GCN and MGCN+SUM, as compared to the gold reference.
The main entity is highlighted in red, while topic-related entities are highlighted in blue.
Given the KG containing all these entities, we intend to generate the description about ``\textit{New Jersey Symphony Orchestra}''.
Firstly, MGCN+SUM is able to cover the main entity and most topic-related entities correctly, while GCN fails to identify the main entity.
This suggests that without multi-graph transformation or effective aggregation methods, it is hard for GCN to extract useful information given a large number of triples in the KG.
Length-wise, the output generated by MGCN+SUM is relatively longer than the one generated by GCN, and thus covers more information. 
We attribute the reason to GCN's deficiency of information loss, as mentioned earlier.


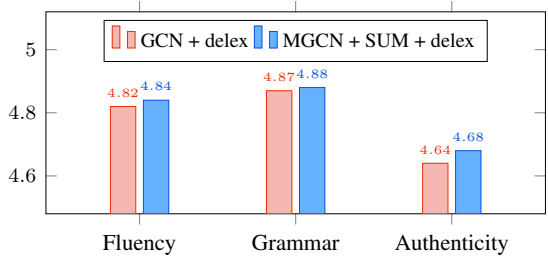
\begin{figure}[t!]
\centering
\scalebox{0.95}{
\begin{tikzpicture}
\pgfplotsset{width=8.5cm,height=4.4cm,compat=1.8}
\begin{axis}[
    ybar,
    enlargelimits=0.3,
    legend style={at={(0.5,0.96)},
      anchor=north,legend columns=-1},
    ybar=3pt,
    ymin=4.6, ymax=5.0,
    symbolic x coords={Fluency,Grammar,Authenticity},
    xtick=data,
    xticklabel style = {font=\fontsize{9}{1}\selectfont},
    yticklabel style = {font=\fontsize{8}{1}\selectfont},
    legend style={font=\fontsize{8}{1}\selectfont},
    nodes near coords,
    nodes near coords align={vertical},
	every node near coord/.append style={font=\fontsize{5}{1}\selectfont}
    ]
\addplot[myred, fill=lowred] coordinates {(Fluency,4.82) (Grammar,4.87) (Authenticity,4.64)};
\addplot[myblue, fill=lowblue] coordinates {(Fluency,4.84) (Grammar,4.88) (Authenticity,4.68)};
\legend{GCN + delex, MGCN + SUM + delex}
\end{axis}
\end{tikzpicture}
}
\caption{\label{fig:human} Results for human evaluation.}
\end{figure}

\paragraph{Human Evaluation}
In order to further assess the quality of the generated sentences, we conduct human evaluation by randomly selecting 100 sentences from outputs generated by GCN+delex and MGCN+SUM+delex.
We hire 6 annotators to evaluate the quality based on three evaluation metrics: fluency, grammar and authenticity.
In terms of authenticity, annotators rate this metric based on the KG (i.e., Wikidata).
More specifically, we give our annotators all main entities' neighbors, 1-hop and 2-hop connections between main entities and topic-related entities as references.
A full score will be given if the statements in the generated sentences are consistent with the facts shown in the KG.
All three metrics take values from 1 to 5, where 5 states the highest score.
The results are shown in Figure \ref{fig:human}.
Recall that BLEU scores of GCN+delex and MGCN+SUM+delex are 28.4 and 30.0 respectively, we can see from Figure \ref{fig:human} that MGCN+SUM+delex only performs slightly better than GCN+delex on the two language quality metrics, namely, fluency and grammar. 
For authenticity, the improvement is more significant. Plausibly it is because the 1.6 BLEU improvement results in more impact on the factual correctness.

\subsection{Additional Experiments}
To examine our model's efficacy on a dataset of different characteristics, we conduct an auxiliary experiment on WebNLG \cite{gardent2017webnlg}, which shares the most similarity with ENT-DESC dataset among those benchmark datasets (e.g., E2E, AGENDA, W\textsc{iki}B\textsc{io}, etc.).
The experiments on WebNLG dataset are under the same settings as the main experiments on our ENT-DESC dataset.

As shown in Table \ref{tab:webnlg}, we observe that our proposed models outperform the state-of-the-art model MELBOURNE.
However, the performance improvement is less obvious on this dataset, largely due to different characteristics between WebNLG and ENT-DESC.
As mentioned in the dataset comparison, the input graphs in WebNLG dataset are much simpler and smaller, where all the information is useful for generation.
Our MGCN model would show stronger advantages when applied to a larger and more complicated dataset (e.g. ENT-DESC dataset),
where extracting more useful entities and relations from the input graphs and effectively aggregating them together is more essential.

\begin{table}[t!]
	\centering
	\scalebox{0.7}{
	    \setlength{\tabcolsep}{2mm}{
        \begin{tabular}{lc}
        \toprule
        \bf Models & \bf BLEU  \\
        \midrule
        TILB-SMT \cite{gardent2017webnlg} & 44.28 \\
        MELBOURNE \cite{gardent2017webnlg} & 45.13 \\
        MGCN & \textbf{45.79}\\
        \midrule
        MGCN $+$ CNN & 45.83 \\
        MGCN $+$ AVG & \textbf{46.55} \\
        MGCN $+$ SUM & 45.23 \\
        \bottomrule
        \end{tabular}}
    }
    \caption{Results on WebNLG dataset.}
	\label{tab:webnlg}
\end{table}

\section{Conclusions and Future Work}
We present a practical task of generating sentences from relevant entities empowered by KG, and construct a large-scale and challenging dataset ENT-DESC to facilitate the study of this task.
Extensive experiments and analysis show the effectiveness of our proposed MGCN model architecture with multiple aggregation methods.
In the future, we will explore more informative generation and
consider applying MGCN to other NLP tasks for better information extraction and aggregation.

\bibliography{emnlp2020}
\bibliographystyle{acl_natbib}

\end{document}